\documentclass[11pt]{article}

\pdfoutput=1

\usepackage{emnlp2022}

\usepackage{times}
\usepackage{latexsym}

\usepackage{multirow}
\usepackage{amsfonts}
\usepackage{graphicx}
\usepackage{pdflscape}
\usepackage{tablefootnote}

\usepackage{float}

\usepackage[T1]{fontenc}

\usepackage[utf8]{inputenc}

\usepackage{microtype}

\usepackage{hyperref}

\newcommand{\squishlist}{
 \begin{list}{$\bullet$}
  { \setlength{\itemsep}{0pt}
     \setlength{\parsep}{3pt}
     \setlength{\topsep}{3pt}
     \setlength{\partopsep}{0pt}
     \setlength{\leftmargin}{1.5em}
     \setlength{\labelwidth}{1em}
     \setlength{\labelsep}{0.5em} } 
}

\newcommand{\squishend}{
  \end{list}  
}

\title{A Pipeline for Generating, Annotating and Employing Synthetic Data for Real World Question Answering}

\author{Matt Maufe \\
  Filament AI, UK\\
  University of Warwick, UK \\ 
  \texttt{matt.maufe@filament.ai} \hspace{1em}\\\And 
  James Ravenscroft \\
  Filament AI, UK\\
  University of Warwick, UK \\
  \texttt{james.ravenscroft@filament.ai} \\\AND
   Rob Procter \\
   University of Warwick, UK \\
   The Alan Turing Institute, UK \\
   \texttt{Rob.Procter@warwick.ac.uk} \\\And
    Maria Liakata \\
  Queen Mary University of London, UK \\
  The Alan Turing Institute, UK \\
  \texttt{m.liakata@qmul.ac.uk} \\
}

\begin{document}
\maketitle

\begin{abstract}
Question Answering (QA) is a growing area of research, often used to facilitate the extraction of information from within documents. State-of-the-art QA models are usually pre-trained on domain-general corpora like Wikipedia and thus tend to struggle on out-of-domain documents without fine-tuning. We demonstrate that synthetic domain-specific datasets can be generated easily using domain-general models, while still providing significant improvements to QA performance. We present two new tools for this task: A flexible pipeline for validating the synthetic QA data and training downstream models on it, and an online interface to facilitate human annotation of this generated data. Using this interface, crowdworkers labelled 1117 synthetic QA pairs, which we then used to fine-tune downstream models and improve domain-specific QA performance by 8.75 F1. 
\end{abstract}

\section{Introduction}
Having enough relevant training data is a key factor for achieving strong performance in machine learning and NLP \cite{chinchilla_data_equation}, but for many tasks, large domain-specific datasets are expensive and time-consuming to create manually. 
This is especially true for tasks like Extractive Question Answering (QA), which both relies on domain-specific knowledge and requires skilled annotators.
These difficulties have led to increased interest in synthetic data generation recently \cite{increased_augmentation_interest} through various methods such as bootstrapping from smaller datasets, or through generative models which create entirely new data.\\\\
We make the following contributions: \newpage
\squishlist
    \item A modular architecture-agnostic pipeline that takes as input unstructured documents and produces both synthetic QA pairs and a QA model trained on them; We show in Section \ref{qa_results} that using this synthetic domain-specific data allows for a dramatic improvement on the QA task compared to baseline state-of-the-art models, especially on unanswerable questions. 
    \item A web-based tool that allows annotators to label various aspects of the synthetic data with ease, alongside guidelines to help ensure consistency and quality in their labels. 
    \item We release\footnote{\href{https://github.com/FilamentAI/qa-annotation}{GitHub}} this annotation tool and its guidelines for general use. While we use and evaluate this pipeline in the domain of business news, the pipeline is sufficiently flexible to be applied to other domains, including potentially being applicable to abstractive QA.

\squishend




\section{Background and Related Work}

\noindent\textbf{Grammaticality Models}
allow for improving the quality of synthetic data and subsequent performance in downstream tasks by better aligning it with real user data. 
On benchmark datasets, such as the Corpus of Linguistic Acceptability (CoLA, \citealp{cola_dataset}) which contains a wide range of examples from published linguistics literature, current state-of-the-art models \cite{ernie} can achieve a Matthew's Correlation Coefficient score \cite{matthews_correlation_coefficient} of approximately 0.775 \cite{cola_sota_mcc}, exceeding human performance (0.713, \citealp{cola_dataset}) in some cases, though this can vary significantly depending on the sentence's syntactic complexity and length \cite{cola_annotated}. \\






\begin{figure*}
    \centering
    \includegraphics[width=0.8\textwidth]{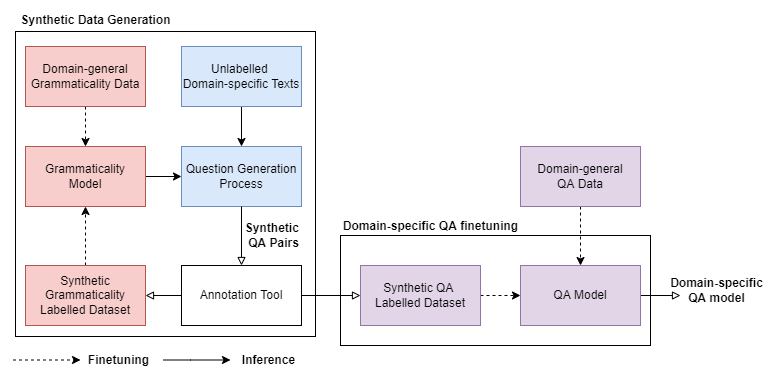}
    \caption{The overall pipeline. The question generation process (blue) generates synthetic QA pairs, which are validated by the grammaticality model. The annotation tool is used to present this data to users for annotation, and the resultant labelled data is then used to fine-tune the grammaticality (red) and QA (purple) models.}
    \label{fig:system_overview}
\end{figure*}

\noindent\textbf{Synthetic NLP Data Generation}
\label{qg_background}
Synthetic data generation is an attractive option for dataset creation, especially for domain-specific tasks. Various methods for bootstrapping from smaller datasets have been devised, such as back-translation \cite{backtranslation} and Sibylvariant transformations \cite{sibylvariant}. Backtranslation produces paraphrases through round-trip translation, while Sibylvariant transformations modify or combine texts in predictable ways to create new data with a different label.


Of particular interest are methods that use text generation models to create entirely new data, rather than simply paraphrasing or combining inputs predictably. A variety of these models have been used to generate new QA pairs \cite{t5_qa}, such as the T5 model \cite{t5} and BERT \cite{bert}. 

Synthetic data generation can be particularly useful when fine-tuning a model on a specific domain, for which manually-curated datasets may not exist. Whilst high quality datasets such as SQuAD 2.0 \cite{squad_2} do exist for QA tasks, they tend to only have general content, e.g. from Wikipedia. 
Thus models trained on them often struggle on more domain-specific tasks (\citealp{domain_transfer}, see also Section \ref{qa_results} below). 

\noindent \textbf{Evaluation of Synthetic QA Pairs} Evaluating Question Generation (QG) models can be difficult due to the nature of the problem: A good question tends to have various qualities (grammatical, answerable, non-trivial to answer, etc.) that are difficult to capture in a single metric, especially one that correlates well with human judgements \cite{qg_evaluation_difficulties}. 
Nonetheless, several metrics such as BLEU \cite{bleu} and BERTScore \cite{bert_score} have been proposed, though they rely on having reference questions available and often do not capture whether or not the question is \emph{answerable} \cite{qg_evaluation_answerability}. However, \citet{squad_2} show that the use of unanswerable questions when training QA models is important for real-world performance, making it a metric of interest.

Round-trip evaluation, such as the methods proposed by \citet{roundtrip_evaluation}, allows for evaluating the generated data by checking how consistent downstream model results are when synthetic data is used as the model input, e.g. if the generated answer is found for a synthetic question when the question is input to a QA model. We adopt this approach and discuss it further in Section \ref{qg_results}. 


\section{System Overview}

\begin{figure*}
    \centering
    \includegraphics[width=0.9\textwidth]{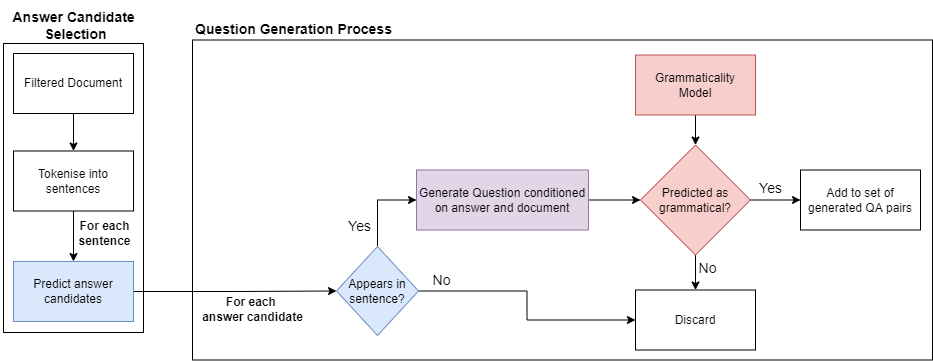}
    \caption{The question generation pipeline.}
    \label{fig:qg_pipeline}
\end{figure*}

Figure \ref{fig:system_overview} shows an overview of our system for creating domain-specific synthetic QA pairs which are used to train downstream models. The QG process (see Section \ref{qg_overview} for details) creates domain-specific QA pairs from unlabelled texts. This data is then annotated for grammaticality and correctness using the annotation tool, allowing for the creation of two new domain-specific datasets to fine-tune both grammaticality and QA models. 

We take a subset of a proprietary knowledge base as our set of input documents and use this to create our domain-specific QA dataset (which we call ``SYFTER''). The knowledge base contains documents obtained by scraping online articles and is focused on business news, such as information about corporate structures, and is thus quite distinct in subject matter from our external domain-general data (SQuAD 2.0, see Section \ref{qg_overview}). 


\subsection{Grammaticality Validation} 
\label{grammaticality_overview}
We use a pre-trained BERT model\footnote{ \href{https://huggingface.co/bert-base-uncased}{bert-base-uncased}} \cite{bert} to evaluate the grammaticality of each synthetic question and answer and we discard ungrammatical ones under the intuition that encouraging the synthetic data to be grammatically correct results in the final dataset being more similar to questions posed by real users and improved performance on the downstream task. 


We use the ``in-domain'' data from the Corpus of Linguistic Acceptability (CoLA, \citealp{cola_dataset}) dataset to train our grammaticality model in the domain-general setting. 



While from a linguistic perspective \cite{grammaticality_gradient}, grammaticality can be seen as either a binary or a gradient feature, we use it as a binary label to better standardise with other papers and with CoLA. Furthermore, annotators are unlikely to hold consistent beliefs about the \emph{degree} to which something is ungrammatical, given the high level of subjectivity inherent in such a judgement, and so treating it as binary reduces the potential for noise in the labels. 

Because both the CoLA and SYFTER grammaticality datasets have a large degree of class imbalance\footnote{Approximately 25\% and 10\% ungrammatical respectively}, we use SMOTE \cite{smote} to oversample the ungrammatical instances and achieve a uniform class distribution.


\subsection{Synthetic Question-Answer Pair Generation}
\label{qg_overview}

The Question Generation process takes as input a natural language document (in our case, a paragraph or a single sentence) and outputs a QA pair that can be answered from this document. This is done using two models: One to select answer candidates from the document, and one that generates a question based on both the answer and the full document, for each candidate. The full process is shown in Figure \ref{fig:qg_pipeline}.

We extend Patil Suraj's question-generation library \cite{patil_suraj} to work with any SQuAD 2.0-format dataset rather than only ones available from \href{https://huggingface.co/}{HuggingFace}, as well as enabling it to gracefully discard invalid answers without breaking, and partially integrating it into our own pipeline. 

We use two separate T5 \cite{t5} models fine-tuned on SQuAD V1\footnote{Due to time constraints, we did not re-train on SQuAD 2.0, but the model performs well nonetheless (Section \ref{qg_results})} data for both answer selection and question generation\footnote{\href{https://huggingface.co/valhalla/t5-small-qa-qg-hl}{valhalla/t5-small-qa-qg-hl} and \href{https://huggingface.co/valhalla/t5-small-qg-hl}{valhalla/t5-base-qg-hl} respectively.}, and specify the task at inference time in natural language following the prompting paradigm \cite{prompting}. We ``highlight'' the answer token during question generation as described in \cite{highlight}.\footnote{E.g. ``generate question: The <hl>dog<hl> is red''.} Because the underlying model is abstractive rather than extractive, it occasionally produces answer candidates that do not appear in the context and are thus unusable for extractive QA, which we discard.

Prior to answer selection, we filter out unsuitable input documents in two stages: We first filter out documents that are very short\footnote{Less than 10 tokens} or which match at least one of a set of RegEx filters (see Appendix \ref{appendix:context_filtering_regex} for details), allowing us to remove any that are clearly semantically null. We then apply a second filter using a BERT Part-of-Speech model\footnote{ \href{https://huggingface.co/vblagoje/bert-english-uncased-finetuned-pos}{vblagoje/bert-english-uncased-finetuned-pos}} such that only documents that contain a verb, or an auxiliary verb and a proper noun, are included so as to remove documents that do not present information that questions can be built around.




Each sentence in each filtered document is input to the answer selection model, which identifies answer candidates within them. Intuitively, a span is an answer candidate if a question can be built around it, and so the model tends to select ones representing entities or relations.

Questions are then generated, conditioned on each answer and the entire associated document, and if validated by the grammaticality model they are added to the synthetic QA dataset.

The resultant dataset can then be input directly into the annotation tool. 

An ablation test over the filters (including the grammaticality model) can be found in Appendix \ref{appendix:qg_filter_ablation}.




\subsection{Question Answering}
\label{qa_overview}

We use an ALBERT \cite{albert} Question Answering model to predict an answer represented as a span within the document, indicated by two token indices (start and end).

The model is able to provide ``null answers'', indicating that the question cannot be answered, either directly or by having its prediction changed to the null answer if the null-answer's confidence score is above a ``null-answer threshold'' (regardless of the original prediction's confidence score).

We utilise SQuAD 2.0 \cite{squad_2} for the initial fine-tuning of our QA model, as it is a large high-quality dataset containing both answerable and unanswerable questions, and as a general-domain dataset it allows us to demonstrate the utility of our domain transfer methods.

The resultant QA model is then fine-tuned on our domain-specific ``SYFTER'' dataset in order to adapt it to our desired domain, which focuses on news articles about commercial events such as product launches and earnings reports (whereas SQuAD's data comes from Wikipedia and focuses more on history, politics, and geography).\footnote{SQuAD's domains can be explored \href{https://rajpurkar.github.io/SQuAD-explorer/explore/v2.0/dev/}{here}.}

\subsubsection{Detecting Unanswerable Questions}
\label{qa_ablation_overview}
During development, we noticed that when trained on a single domain (SQuAD or SYFTER), the QA models could learn to effectively identify if a question from that domain could be answered or not, but performance on this task would drop significantly when trained on both domains.

This was likely due to a combination of our ``unanswerable question'' label being applied more broadly (to nonsensical questions as well as unanswerable ones), and due to the significant amount of class imbalance in the dataset (especially for the SYFTER data), as well as a small amount of noise in the labels detected through manual inspection.

    
    
    
We explored various methods to resolve this problem when using combined training data, and discuss an ablation study over them in Appendix \ref{appendix:ablation}, with results in Table \ref{table:ablation_results}. 

\squishlist
\item We appended ``source markers'' to the end of each question, prior to tokenisation, which indicated the domain that the question came from: either ``[SQuAD]'' or ``[SYFTER]'', in order to allow the model to better learn domain-specific features. 

\item We tuned the `null-answer threshold'' on the validation set.

\item We investigated training the model simultaneously for the tasks of both QA and sequence classification as ``answerable'' / ``unanswerable''. This follows findings from \citet{multitask_survey} that multitask learning can often improve performance, and given the interdependence between question answering and detecting if a question \emph{can} be answered.

\item Finally, we used alpha-weighted Focal Loss \cite{focal_loss} rather than Cross Entropy Loss for sequence classification in the multitask setting to better handle class imbalance. 
\squishend


\subsection{Data Annotation}
\label{annotation_overview}

\begin{figure*}
    \centering
    \includegraphics[width=0.8\textwidth]{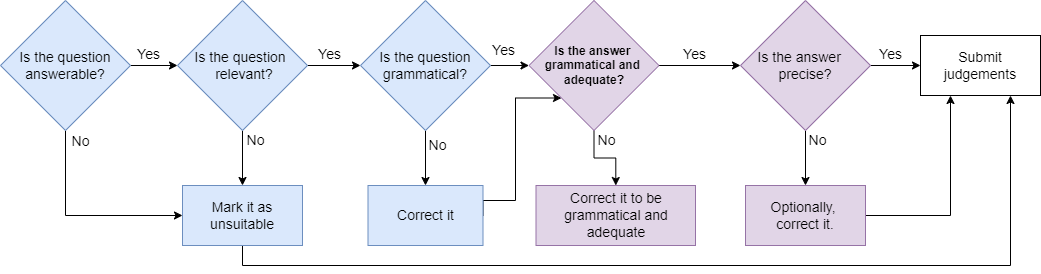}
    \caption{The annotation process. The answerability and relevance of questions (blue) is dependent on the document, without considering external knowledge. Answers (purple) must appear within the document to be accepted.}
    \label{fig:qa_annotation_process}
\end{figure*}

\begin{table*}
\begin{center}
\begin{tabular}{c|c|c|c}
     Model & Training Data & \# Train Examples & Macro F1 Score \\
     \hline
     BERT & CoLA & 10584 & 61.18 \\
     BERT & SYFTER & 2796 & \textbf{75.74} \\
     \hline
     BERT & CoLA + SYFTER & 13608 & 74.68 \\
\end{tabular}
\end{center}
\caption{The Grammaticality model results. The best setting is indicated in \textbf{bold} text. ``\# Train Examples'' refers to the data \emph{after} oversampling.}
\label{table:grammaticality_results}
\end{table*}

In order to label the synthetic data for supervised training, we created an annotation tool\footnote{\href{https://www.youtube.com/watch?v=VHBHE1pVWzA}{A video demo can be found here}} using Streamlit \cite{streamlit} which allows annotators to view model-generated QA pairs, along with their associated context document, and annotate them in various ways. An example of how QA pairs are presented within the tool can be found in Figure \ref{fig:qa_annotation_tool} in Appendix \ref{appendix:annotation_tool}.

We used a series of three preliminary studies to iteratively refine our annotation tool and guidelines, with each study involving 10 participants (who did not participate in subsequent studies). This allowed us to identify and fix any points of misunderstanding before using the tool for the final annotation study on the entire dataset. As with the final annotation study, these were done via Prolific\footnote{\url{https://www.prolific.co/}}
 and under the same annotator filters (as well as filtering out previous participants).

Following each preliminary study, we followed up with annotators in cases where they had made unintuitive judgements or appeared to have misunderstood, and used these discussions to refine the guidelines presented. The final guidelines are shown in Appendix \ref{appendix:annotation_guidelines}. 


Each annotator was assigned to a group with two others, and each group of three annotators provided annotations for 2\% of the total dataset, with gold labels coming from majority judgements.

The annotation process is shown in Figure \ref{fig:qa_annotation_process}. Questions marked as unsuitable (for either reason) are not labelled further, and comprise the set of unanswerable questions for the SYFTER domain.

Questions were judged on suitability (whether the question is answerable and relevant to the document) as well as grammaticality.

Grammaticality for both questions and answers was posed to annotators as a question of ``reading naturally'', in order to better mimic real user questions and avoid the subjective issues inherent to judging grammaticality. 

Answers were judged on both naturalness and quality. In the latter case, an answer was considered ``adequate'' if it answered the question but had either extraneous details or was missing details, and ``precise and correct'' if it answered the question with all of the relevant details, but no more.

We asked annotators to rewrite questions and answers that did not read naturally, as well as inadequate answers, and did not allow for the submission of the labels until the texts were corrected or the question was marked as unsuitable (e.g. if they could not be corrected within our constraints).

\section{Experiments and Results}
\label{experiments_and_results}

\begin{table*}
    \begin{center}
        \begin{tabular}{c|c|c|c|c}
            Test Dataset & QA Model & Exact Match & Similarity\\
             \hline
             SQuAD 2.0 & RoBERTa & \textbf{67.81\%} & \textbf{81.89\%} \\
             SYFTER & RoBERTa & 64.55\% & 77.27\%
        \end{tabular}
    \end{center}
    \caption{Roundtrip evaluation of our QA datasets' quality, using an off-the-shelf QA model. The RoBERTa model was trained on SQuAD 2.0. Best results indicated in bold text.}
    \label{table:qg_results}
\end{table*}

\begin{table*}
\begin{center}
    \begin{tabular}{c|c|c}
        Document & Question & Answer \\
        \hline 
        \multirow{3}{0.5\textwidth}{"International law firm Ashurst announces the appointment of Matthias Weissinger as partner in Munich.} & \multirow{3}{0.25\textwidth}{Who is the new partner of Ashurst in Munich?} & Matthias Weissinger \\
        & & \\
        & & \\
        \hline
        \multirow{4}{0.5\textwidth}{To date we've delivered more than one billion pieces of protective equipment to the frontline.} & \multirow{4}{0.25\textwidth}{How many pieces of protective equipment have been delivered to the frontline?} & more than one billion \\
        & & \\
        & & \\
        & & \\
        \hline
        \multirow{2}{0.5\textwidth}{As a major food sector player, Bel fully assumes its duty to do everything possible to ensure the continuity of its operations.} & \multirow{2}{0.25\textwidth}{What sector is Bel a major player in?} & food \\
        & & \\
    \end{tabular}
\end{center}
\caption{Example Question-Answer Pairs Generated from Documents}
\label{table:example_qa_pairs}
\end{table*}


\begin{figure*}
    \centering
    \includegraphics[width=0.8\textwidth]{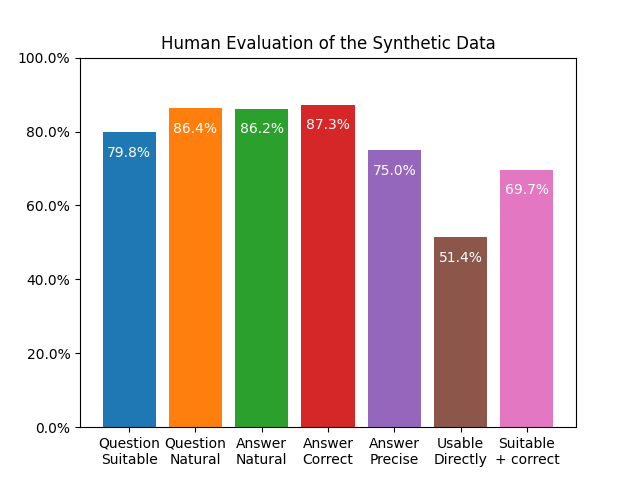}
    \caption{Human Evaluation results on the annotated data. Only QA pairs that had a suitable question were judged further on the other metrics. Percentages shown are based on annotator consensus rather than individual judgements.}
    \label{fig:human_qg_evaluation}
\end{figure*}

\begin{table*}
\begin{center}
\begin{tabular}{c|c|c|c|c|c|c|c|c}
    Model & Training Data & \% Synthetic & \multicolumn{2}{c}{Answerable} & Unanswerable & \multicolumn{2}{c}{Overall} \\
      & & Training Data & EM & F1 & EM & EM & F1 \\
     \hline
     ALBERT & SQuAD 2.0 & 0\% & \textbf{84.87} & \textbf{91.09} & 12.16 & 61.06 & 65.25 \\
     ALBERT & SYFTER & 100\% & 53.26 & 59.71 & \textbf{72.00} & 57.26 & 63.34 \\
     \hline
     ALBERT & SQuAD 2.0 + SYFTER & 0.62\% & 71.74 & 83.24 & 40.00 & \textbf{64.96} & \textbf{74.00} \\
\end{tabular}
\end{center}
\caption{Question Answering model results on the SYFTER test set. The best settings are shown in \textbf{bold}.}
\label{table:qa_results}
\end{table*}

The Grammaticality and Question Answering models are tested in both the setting of interest (combined domain-general and domain-specific data) as well as two baseline data settings (domain-general data only\footnote{CoLA for the grammaticality task, SQuAD for the QA task}, and domain-specific data only). This allows us to both measure how useful the synthetic data is as an addition to domain-general data and to also evaluate the feasibility of fine-tuning using \emph{only} synthetic data, which would reduce time and expense significantly given its small size. 

The combined test sets for the Grammaticality and QA models are produced by combining the appropriate domain-general data (CoLA or SQuAD) with the domain-specific SYFTER data and then testing the model on this combination dataset. 


We evaluate the Question Generation process in both the domain-general and domain-specific settings, but do \emph{not} evaluate the combined setting due to the nature of the evaluation (see Section \ref{qg_results}).






\subsection{Grammaticality Classification}
\label{grammaticality_results}
We evaluate the grammaticality model using the model's F1 score, treating grammaticality as a binary sequence classification task, and achieve strong results in both the synthetic-only and combined data settings, as shown in Table \ref{table:grammaticality_results}. The domain-specific model actually performs better than both the domain-general model and the combined-data setting, despite training on only a small amount of synthetic data, indicating the importance of using domain-specific data during training. 




\subsection{Synthetic Question-Answer Pair Generation}
\label{qg_results}

We evaluate the synthetic questions through roundtrip evaluation as discussed in Section \ref{qg_background}. For each generated QA pair, we use an off-the-shelf QA model\footnote{\href{https://huggingface.co/deepset/roberta-base-squad2}{deepset/roberta-base-squad2}, which has strong performance on SQuAD 2 data} to answer the generated question (based on its associated context) and then compare the answers in two ways: Exact match; and comparing their similarity with their most-similar question at the token level using length-normalised Levenshtein distance \cite{levenshtein} via NLTK \cite{nltk}. Intuitively, if the question is well-formed and precise, and the answer is relevant to it, the QA model should find the correct answer. 

As shown in Table \ref{table:qg_results}, the synthetic data is of high quality, reaching similar levels to SQuAD 2.0, which was manually created by humans. Furthermore, Table \ref{table:example_qa_pairs} shows examples of the synthetic data produced and used. The generated questions are both fluent and of interest, and the answers are both precise and correct. The first question is slightly stilted, but still easily understandable.

Finally, the annotation process can also be thought of as a form of human evaluation and, as shown by Figure \ref{fig:human_qg_evaluation}, the vast majority of the data was found to be of high-quality (suitable, reading naturally, and correct+precise answers). However,  48.6\% of the data, including unsuitable questions, did require some input from annotations in some form (not counting data that was imprecise but otherwise good). This indicates that while the data tends to be of high-quality overall, about half of the datapoints do contain a small amount of noise. 69.7\% of the questions are suitable and have correct answers, which can be considered the key factors for good synthetic QA data, and as such a high percentage of the data could be used to train a QA system as-is without needing corrections.

\subsection{Question Answering}
\label{qa_results}

We take approximately 11.6\% of the total annotated SYFTER data (117 questions, approximately 21\% of which are unanswerable) to use as the QA test set, and split it at the document-level to avoid potential information leaks from the training data. 


The QA model is evaluated through both the ``Exact Match'' (EM) score, and at the token level using F1 score, via the \href{https://github.com/huggingface/datasets/blob/2.4.0/datasets/squad_v2/squad_v2.py}{HuggingFace wrapper} around the official SQuAD evaluation script. In both cases, the text is first lowercased and normalised to remove articles and standardise whitespace. EM and F1 are identical for unanswerable questions.

We present the results from the best setting, which uses null-answer threshold tuning and multitask learning \emph{without} Focal Loss (see Appendix \ref{appendix:ablation}), in Table \ref{table:qa_results}.

The SYFTER-only model performs well despite the SYFTER dataset being much smaller than SQuAD 2.0, and is much better at handling unanswerable questions. By combining the two, we achieve the best overall performance, and maintain reasonable performance on unanswerable questions despite the issues discussed in Section \ref{qa_ablation_overview}.


\section{Conclusion}
We present a pipeline for using and evaluating synthetic QA data and an interface for annotating it, as well as annotation guidelines. The combination of domain-general and synthetic data allows our QA model to perform significantly better (+ 9 F1) on domain-specific documents than it did when trained solely on a similar amount of domain-general data. The pipeline is simple to apply to both current and future state-of-the-art models, enabling better performance in low-resource domains. 

\section{Acknowledgements}
This work is supported by an Innovate UK grant under the KTP scheme (KTP 11714) to Rob Procter, Maria Liakata and Filament AI, as well as the Alan Turing Institute (EP/N510129/1). Maria Liakata has been supported by a UKRI/EPSRC Turing AI Fellowship EP/V030302/1. 

\section{Limitations}

Whilst our system demonstrates that we can achieve significant improvements from synthetic domain-specific data with minimal additional time and expense, it does have certain limitations: We do not consider ``adversarial questions'' when training, and it thus would likely struggle on these kinds of questions based on findings such as those from \citet{adversarial_qa}. 

We also found that our synthetic data primarily consists of questions which identify entities (e.g. ``Who is the CEO of Microsoft?'', ``When did Microsoft acquire Bethesda Softworks?'', ``What are the five principles of good leadership?''), and does not contain many examples of questions about relationships between entities (e.g. ``Is selling ice cream more profitable than selling widgets?''), and answers to the latter may be of relatively poor quality. 

This is likely due to what appears to be a similar trend in SQuAD V1 that the Question Generation model was trained on: SQuAD primarily asks questions with short entity-focused answers (dates, names, etc.) \cite{squad_question_types} and approximately half of the answers in SquAD \cite{squad_1} are proper nouns, dates, or other numbers indicating that their corresponding questions are likely entity-focused.

The questions of interest to us are generally entity-based and so this limitation does not directly impact our own usage of the model, but we recognise that it potentially limits its applicability to other domains. In the future, the model's performance on non-entity questions could be investigated and improved through tools like AdaTest \cite{adatest}.

The tool also still requires some amount of human involvement to annotate and filter the synthetic data, and the Grammaticality model results (Table \ref{table:grammaticality_results} indicates that filtering with purely domain-general models would be ineffective. However, it is possible to generate the QA pairs without annotation and, given the high quality of the data (Figure \ref{fig:human_qg_evaluation}), it may be reasonably possible to use the data directly (treating it all as suitable and grammatical) to achieve a still-significant boost to domain-specific performance.

The main problem with not using human annotation would be that our ``unanswerable questions'' are all ones marked as ``unsuitable'' by humans, and thus using the synthetic data directly would lead to only having synthetic questions that are considered to be answerable. This could be improved through extending the QG pipeline to also produce deliberately-unanswerable examples, but is not currently possible. 

Finally, whilst we use the grammaticality model for validation during the question generation process, we do not train either the Answer Selection or Question Generation models with grammaticality as a second objective function. Training it in a multitask setting would likely have guided it towards producing better input, and may have produced more (valid) data from the corpus.

\section{Ethics Statement}

Machine learning tasks often involve the potential for ethical issues, especially when using human annotators to label data. We chose to use Prolific\footnote{\href{https://www.prolific.co/}{https://www.prolific.co/}} as a platform to find and pay annotators, as it offered a reputation for enforcing ethical payments as well as useful filters such as education level and native language. 

We also submitted our project to the University of Warwick's internal ethics process, and were approved without having to make any adjustments.

Prolific annotators are paid a fixed amount, but if a task's average hourly payment falls below a minimum (£5 / \$6.50 per hour), it is required to rectify this and increase the payments.

The mean rate of pay for annotators was reported as £15.63 during the preliminary studies and £15.50 during the primary annotation study, though these figures are \emph{under-estimates} as our own time-tracking indicates that annotators generally spent a significant amount of time not annotating the data questions (but still recorded by Prolific as being on-task). This is well in excess of the UK living wage of £9.50, as well as the ``real living wage'' of up to £11.05 proposed by The Living Wage Foundation\footnote{As discussed \href{https://www.livingwage.org.uk/}{here}.}.

The use of synthetic data does have some inherent potential ethical issues: ``Model hallucination'' is a well-known phenomenon where models can create unfaithful data (e.g. convincing, but false answers to questions) and which can cause real-world harm if the information it provides is acted on \cite{hallucination}. This can affect our own models if the data generation models hallucinate and lead to the QA model internalising incorrect knowledge.

Thankfully, there are various ways to identify these occurrences and mitigate this harm, including perhaps the simplest method of specifying the context in which the data was created and used at appropriate downstream points, so that users can better assess its veracity for themselves. 

To limit this harm, we strongly suggest that other researchers take this into account in their own work, and take the appropriate actions, for instance using human annotators to verify the data and actively designing models to be robust against hallucination, as done in work like \citet{anti_hallucination}.

Finally, despite using a model to create our QA data, and the fact that synthetic data can clearly be very useful, bias is still likely to exist in the data (carried forward from both the model's original training data and the human factor of the annotation done), and we suggest that any data produced be investigated and debiased through tools like AdaTest \cite{adatest}.

\bibliography{custom}
\bibliographystyle{acl_natbib}

\clearpage

\appendix

\section{RegEx Document Filters}
\label{appendix:context_filtering_regex}
Table \ref{table:regex_document_filters} shows the different RegEx filters that we apply to documents in order to filter out ones that are likely to be difficult to select valid answers from. Documents are filtered if \emph{any} substring in them is a match for the expression.

The first expression, which filters out documents that appear to be too similar to contracts, additionally contains certain \emph{whitelist} expressions which prevent otherwise-matching documents from being removed. These can be seen in Table \ref{table:regex_document_filters_whitelist}. In order to be whitelisted, the text that matched the initial filter must fully match the whitelist expression (though the entire document does not have to match).

For clarity when dealing with leading/trailing whitespace, each expression is wrapped in ``double quotes'', but these quotes are not part of the actual expression. Matches with each expression are \textbf{emphasised} for clarity.

\begin{table*}
\begin{flushleft}
\begin{center}
    \begin{tabular}{p{0.6\textwidth-2\tabcolsep}|p{0.2\textwidth-2\tabcolsep}|p{0.2\textwidth-\tabcolsep}}
        RegEx Expression & Intended Matches & Example Match \\
        \hline 
        ``\verb| ?\([0-9A-Za-z]+\)(\([0-9A-Za-z]+\))*|'' & Contract-like documents & ``B 1: Financial Instruments according to Regulation 17\textbf{(1)(a)} of the Regulations''\\
        \hline
        ``\verb|^[0-9]+\.? ?.+|'' & Numeric List & ``\textbf{1. Reassure customers and employees}'' \\
        \hline
        ``\verb|^[ivx]+\.? .+|'' & Roman-numeric List & ``\textbf{xi If the financial instrument has such a period}'' \\
        \hline
        ``\verb|\[ ?\]|'' & Empty square brackets & ``\textbf{[ ]} An acquisition or disposal of financial instruments'' \\
        \hline
        ``\verb|Regulation(s)? [0-9]+|'' & Regulations contract-like & ``B 2: Financial Instruments with similar economic effect according to \textbf{Regulation 17} of the Regulations'' \\
        \hline
        ``\verb|^.{0,15}$|'' & Very short documents & ``\textbf{content}'' \\
        \hline
        ``\verb|^(.{0.5})?\(.+\).{0,5}$|'' & Mostly in brackets & ``\textbf{(please tick the appropriate box or boxes):}'' \\
    \end{tabular}
\end{center}
\end{flushleft}
\caption{RegEx Filters for Documents}
\label{table:regex_document_filters}
\end{table*}

\begin{table*}
\begin{center}
    \begin{tabular}{p{6cm}|p{3cm}|p{6cm}}
        RegEx Expression & Purpose & Example Documents Whitelisted \\
        \hline 
        ``\verb| ?\([A-Z]+s?\)|'' & Allow acronyms & ``CPE Lite is Huawei's latest mini customer premises equipment\textbf{ (CPE)}.'' \\
        \hline
        ``\verb| ?\([A-Z]?[0-9a-z]{4,}\)|'' & Allow short bracketed words & ``Bel reported strong sales momentum in the first two months of the year in global\textbf{(mature)} markets''
    \end{tabular}
\end{center}
\caption{RegEx Whitelists for Documents, applied to the ``Contract-like'' filter.}
\label{table:regex_document_filters_whitelist}
\end{table*}

\clearpage

\section{Question Answering Ablation}
\label{appendix:ablation}
We performed an ablation study over the Question Answering Model components discussed in Section \ref{qa_ablation_overview}, and found that in some cases they significantly improve the performance on unanswerable questions, especially the use of multitask learning. The results of this ablation are shown in Table \ref{table:ablation_results}. 

Whilst we found that some settings (Source Markers, Focal Loss) did not appear to be useful, we nonetheless believe that the utility of source markers when using more domains would be an interesting avenue for future investigation.

\begin{table*}
\begin{center}
\begin{tabular}{c|c|c|c|c|c|c}
     Source Markers & Threshold Tuning & Multitask & Focal Loss & \multicolumn{3}{c}{Performance Gain (F1)} \\
     & & & & Answerable & No Answer & Overall \\
     \hline
     x & x & x & x & 91.02 & 72.97 & 85.11 \\
     \hline
     \checkmark & x & x & x & - 2.2 & + 0 & - 1.48 \\
     x & \checkmark & x & x & - 0.66 & + 1.35 & + 0 \\
     x & x & \checkmark & x & \textbf{- 1.98} & \textbf{+ 4.06} & \textbf{+ 0} \\
     x & x & \checkmark & \checkmark & - 2.14 & + 0 & - 1.44 \\
     \hline
     \checkmark & \checkmark & \checkmark & \checkmark & - 1.91 & + 1.35 & - 0.84 \\
\end{tabular}
\end{center}
\caption{Relative performance gains on the ALBERT QA model in different training settings. A checkmark indicates that the component was used, an ``x'' that it was not. Focal loss is only applicable in the multitask setting. Best setting shown in \textbf{bold}.}
\label{table:ablation_results}
\end{table*}

\section{Question Generation Filter Ablation}
\label{appendix:qg_filter_ablation}
We performed an ablation study over the Question Generation filters discussed in Section \ref{qg_overview} and found that the individual filters tend to have a significant impact on the model's performance on unanswerable questions, but relatively little when considering answerable questions. Given that the filters were primarily designed to filter out documents that were likely to produce low-quality unanswerable questions, this is as expected. The set of filters that we used does not provide the best \emph{overall} F1 Score, but provides a model whose performance is significantly more balanced than the nominally best-performing model, a trait that we found valuable. 

For these tests, we trained and tested the QA model \emph{only} on SYFTER data so as to most clearly see the effects of the filter(s) used (since SQuAD data is not filtered in our pipeline). 

\begin{table*}
\begin{center}
\begin{tabular}{c|c|c|c|c|c|c}
     \multicolumn{4}{c}{Filter} & \multicolumn{3}{c}{Performance Gain (F1)} \\
     Length & RegEx & Part of Speech & Grammaticality & Answerable & No Answer & Overall \\
     \hline
     x & x & x & x & 72.35 & 40.00 & 66.22 \\
     \hline
     \checkmark & x & x & x & 65.60 & 48.00 & 62.16 \\
     x & \checkmark & x & x & 73.8 & 24 & 64 \\
     x & x & \checkmark & x & \textbf{73.67} & 52 & \textbf{69.5} \\
     x & x & x & \checkmark & 71.95 & 36 & 65.24 \\
     \hline
     \checkmark & \checkmark & \checkmark & \checkmark & 59.71 & \textbf{72.00} & 63.34 \\
\end{tabular}
\end{center}
\caption{Relative QA performance gains on the SYFTER test set model using different SYFTER training data filtered in different ways. A checkmark indicates that the component was used, an ``x'' that it was not. Best setting shown in \textbf{bold}. Only SYFTER data was used for training.}
\label{table:qg_filter_ablation_results}
\end{table*}

\clearpage

\section{Annotation Tool}
\begin{figure*}
    \includegraphics[width=\textwidth,height=\textheight,keepaspectratio]{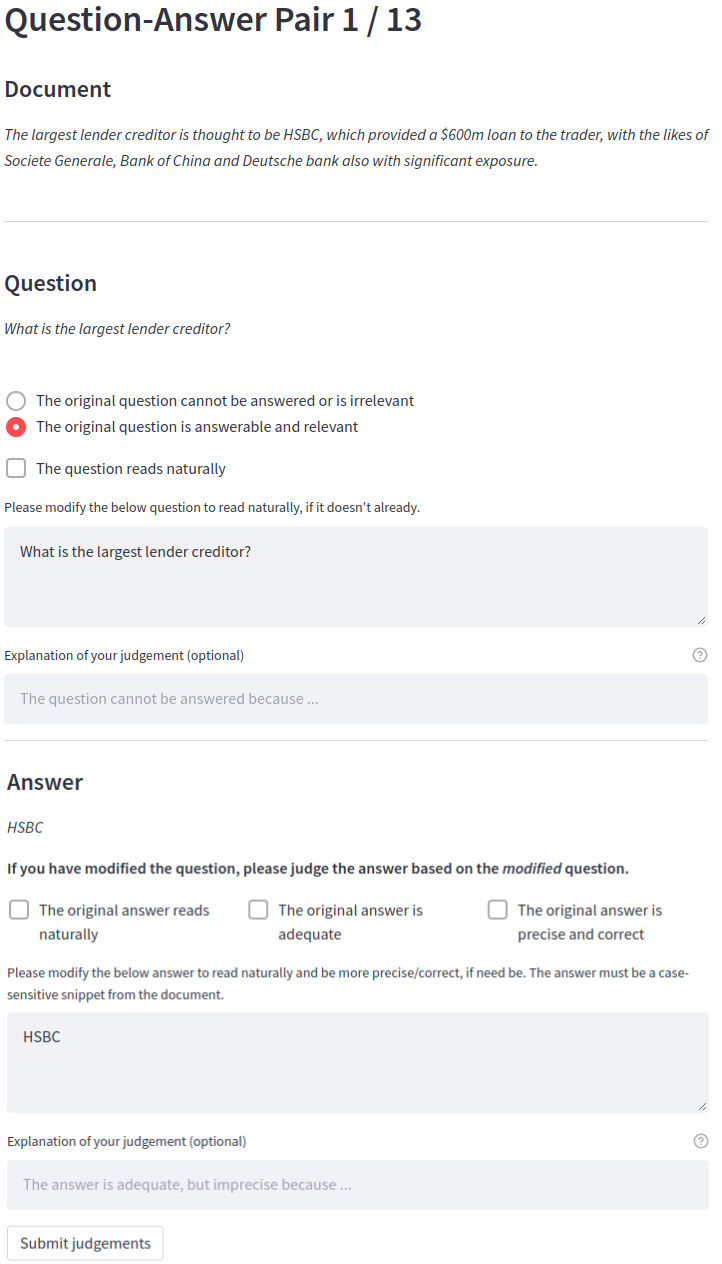}
    \caption{An example of how QA pairs are presented in the annotation tool.}
    \label{fig:qa_annotation_tool}
\end{figure*}

Figure \ref{fig:qa_annotation_tool} shows an example of how QA Pairs are presented to annotators in the annotation tool. See Section \ref{annotation_overview} for details.

A video demo of the tool can be found \href{https://www.youtube.com/watch?v=VHBHE1pVWzA}{here}

\label{appendix:annotation_tool}

\subsection{Annotation Guidelines}
\label{appendix:annotation_guidelines}
\begin{figure*}
    \includegraphics[width=\textwidth,height=\textheight,keepaspectratio]{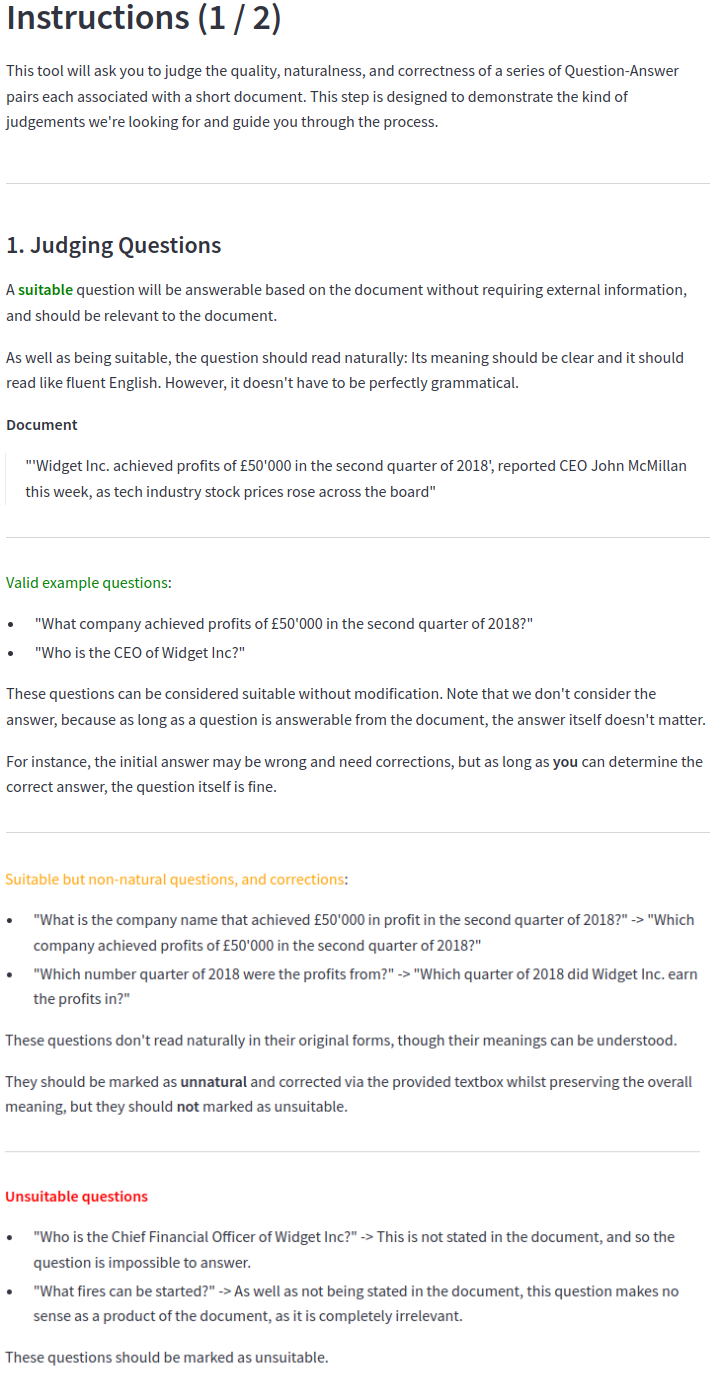}
    \caption{Annotation guidelines for judging question suitability and naturalness.}
    \label{fig:qa_annotation_instructions_questions}
\end{figure*}

\begin{figure*}
    \includegraphics[width=\textwidth,height=\textheight,keepaspectratio]{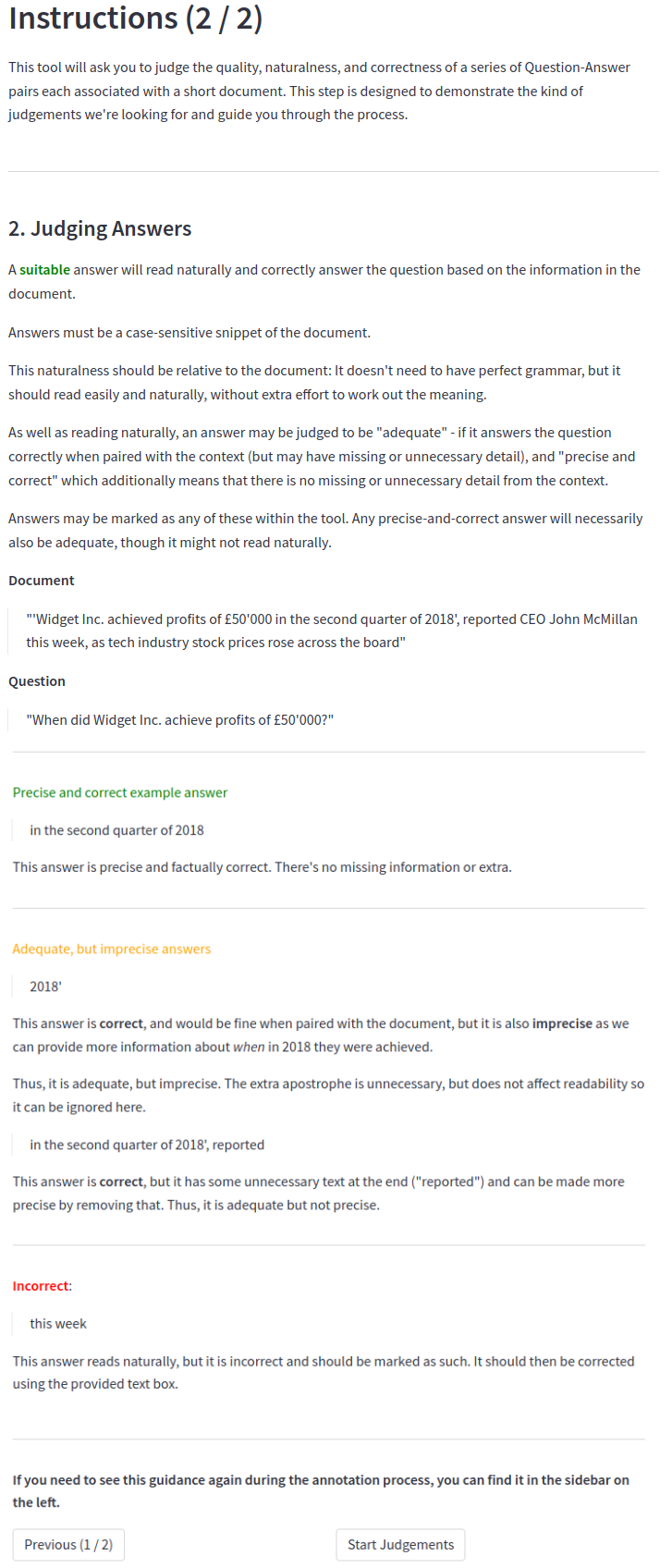}
    \caption{Annotation guidelines for judging answer naturalness and quality.}
    \label{fig:qa_annotation_instructions_answers}
\end{figure*}

We present a set of annotation guidelines which can be given to annotators in order to obtain consistent labels by ``calibrating'' their expectations of what is and is not a valid QA pair. The guidelines for labelling questions can be found in Figure \ref{fig:qa_annotation_instructions_questions} and for answers in Figure \ref{fig:qa_annotation_instructions_answers}.

\end{document}